%% file: main.tex
\documentclass[10pt,twocolumn,letterpaper]{article}

\usepackage[]{cvpr}      


\definecolor{cvprblue}{rgb}{0.21,0.49,0.74}
\usepackage[pagebackref,breaklinks,colorlinks,allcolors=cvprblue]{hyperref}

\usepackage{epigraph}
\usepackage{booktabs}
\usepackage{xcolor}

\usepackage{graphicx}
\usepackage[table,xcdraw,dvipsnames]{xcolor}
\usepackage{multirow}
\usepackage{pifont}
\usepackage{colortbl}
\usepackage{booktabs}
\usepackage{tcolorbox}
\usepackage{soul}
\usepackage{fontawesome} 
\usepackage{graphicx}
\usepackage{enumitem}
\usepackage{dsfont}
\usepackage{multirow}
\usepackage[ruled,linesnumbered]{algorithm2e}

\usepackage{newfloat}
\usepackage{listings}
\DeclareCaptionStyle{ruled}{labelfont=normalfont,labelsep=colon,strut=off} 
\usepackage{xcolor}
\usepackage{dsfont}  

\usepackage{array}
\usepackage{ragged2e}
\usepackage{graphicx}

\usepackage{tabularx}
\def\paperID{10550} 
\def\confName{CVPR}
\def\confYear{2026}
\newcommand\blfootnote[1]{%
  \begingroup
  \renewcommand\thefootnote{}\footnote{#1}%
  \addtocounter{footnote}{-1}%
  \endgroup
}

\title{One Size, Many Fits: Aligning Diverse Group-Wise Click Preferences in Large-Scale Advertising Image Generation}

\author{
    Shuo Lu$^{\spadesuit, \clubsuit, \star}$, Haohan Wang$^{\diamondsuit, \star}$, Wei Feng$^{\diamondsuit, \ddagger}$, Weizhen Wang$^\diamondsuit$, Shen Zhang$^\diamondsuit$, Yaoyu Li$^\diamondsuit$, \\
    Ao Ma$^\diamondsuit$, Zheng Zhang$^\diamondsuit$, Jingjing Lv$^\diamondsuit$, Junjie Shen$^\diamondsuit$, Ching Law$^\diamondsuit$, Bing Zhan$^{\spadesuit}$, \\
    Yuan Xu$^{\spadesuit}$, Huizai Yao$^{\heartsuit}$, Yongcan Yu$^{\spadesuit}$, Chenyang Si$^{\triangle}$, Jian Liang$^{\spadesuit, \clubsuit, \dagger}$ \\[2mm]
    $^\spadesuit$NLPR \& MAIS, CASIA \quad
    $^\clubsuit$School of AI, UCAS \quad
    $^\diamondsuit$JD.COM \quad
    $^\heartsuit$HKUST(gz) \quad
    $^\triangle$PRLab, NJU \\[2mm]
}

\begin{document}
\maketitle

\blfootnote{%
    \hspace{-2em} 
    $^\star$Equal contribution \quad
    $^\ddagger$Project Leader \\
    $^\dagger$Corresponding author: {\tt\small liangjian92@gmail.com}
}

\input{sec/0_abstract}    
\input{sec/1_intro}
\input{sec/2_related_work}
\input{sec/3_method}

\input{sec/4_dataset}
\input{sec/5_experiment}

\input{sec/6_conclusion}

{
    \small
    \bibliographystyle{ieeenat_fullname}
    \bibliography{main}
}
\input{sec/7_suppl}

\end{document}

%% file: sec/0_abstract.tex
\vspace{-10pt}
\begin{abstract}
Advertising image generation has increasingly focused on online metrics like Click-Through Rate (CTR), yet existing approaches adopt a ``one-size-fits-all" strategy that optimizes for overall CTR while neglecting preference diversity among user groups. 
This leads to suboptimal performance for specific groups, limiting targeted marketing effectiveness.
To bridge this gap, we present \textit{One Size, Many Fits} (OSMF), a unified framework that aligns diverse group-wise click preferences in large-scale advertising image generation. 
OSMF begins with product-aware adaptive grouping, which dynamically organizes users based on their attributes and product characteristics, representing each group with rich collective preference features.
Building on these groups, preference-conditioned image generation employs a Group-aware Multimodal Large Language Model (G-MLLM) to generate tailored images for each group. The G-MLLM is pre-trained to simultaneously comprehend group features and generate advertising images.
Subsequently, we fine-tune the G-MLLM using our proposed Group-DPO for group-wise preference alignment, which effectively enhances each group's CTR on the generated images.
To further advance this field, we introduce the Grouped Advertising Image Preference Dataset (GAIP), the first large-scale public dataset of group-wise image preferences, including around 600K groups built from 40M users.
Extensive experiments demonstrate that our framework achieves the state-of-the-art performance in both offline and online settings. 
Our code and datasets will be released at \url{https://github.com/JD-GenX/OSMF}.
\end{abstract}

%% file: sec/1_intro.tex
\vspace{-21pt}
\section{Introduction}
\vspace{-10pt}
\label{sec:intro}
\medskip  
\noindent{\normalsize\itshape Diversity is the one true thing we all have in common.}\\[1ex]
\makebox[\linewidth][r]{\normalsize\itshape \textit{---}Winston Churchill}
\\[0.5ex]

\begin{figure*}
    \centering
    \includegraphics[width=1\linewidth]{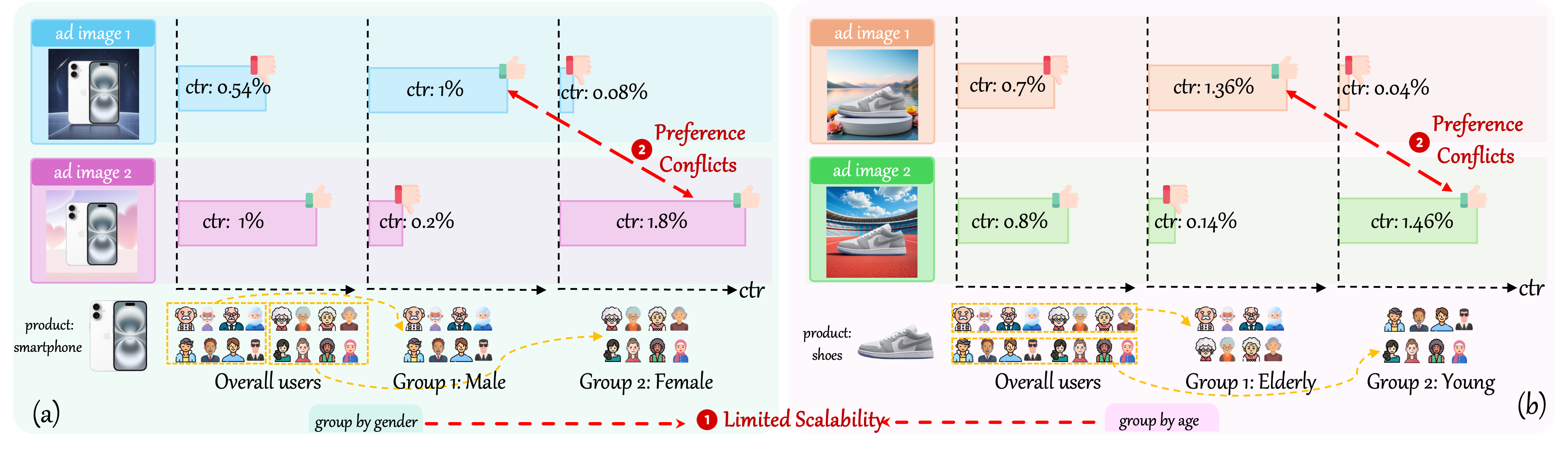}
    \caption{Current methods use overall CTR as unified preference targets, failing to capture group-specific preferences. (a) In smartphone advertising, male and female users show distinct preferences; (b) In shoe advertising, young and old users exhibit distinct preferences. These image preferences are often conflicting in both cases.}
    \label{fig:figure1}
\end{figure*}

Advertising image generation is the process of creating visually compelling content for products, essential for customer attraction and sales growth.
Recent advances~\cite{chen2025ctr, shopee} have shifted towards incorporating Click-Through Rate (CTR) as the reward signal in Reinforcement Learning from Human Feedback (RLHF), marking a promising step towards aligning with online user behavior. 
Despite significant progress, existing methods rely on a flawed assumption: treating overall user preferences as universally applicable. 
However, on popular e-commerce platforms with tens of millions of users, preferences vary substantially across user groups. 
This \textit{one-size-fits-all} approach neglects preference diversity, where even images with high overall CTR may still fail to engage specific user groups, 
thereby limiting targeted marketing effectiveness and generating commercially ineffective assets that squander advertising spend while undermining user engagement.

In light of this limitation, we aim to align diverse group-wise click preferences in large-scale advertising image generation. In this context, group-wise RLHF approaches~\citep{ramesh2024group, zhang2024diverging, park2024rlhf} show promise, which typically involve grouping users into distinct groups and subsequently performing preference alignment for each.
However, this paradigm faces two significant challenges in industrial-scale advertising scenarios. 
In what follows, we elaborate on these challenges and present \textit{One Size, Many Fits} (OSMF), a unified framework to align diverse group-wise click preferences effectively.

\textbf{Limited Scalability of User Grouping.} Existing methods~\citep{ramesh2024group, chakraborty2024maxmin, li2024personalized} employ fixed user attributes to aggregate users for subsequent preference alignment. However, these rigid groupings often fail to adapt across diverse product categories. 
For instance, Figure~\ref{fig:figure1}~(a) illustrates that gender is a primary preference differentiator for smartphone images, whereas in Figure~\ref{fig:figure1} (b) for shoes, age-based distinctions are more pronounced.
To address this challenge, we propose Product-Aware Adaptive Grouping (PAAG). 
It dynamically models user preferences by integrating user attributes (e.g., gender, age, location) with product characteristics, 
producing product-aware user preference features.
PAAG then adaptively clusters users based on these features and aggregates them into rich collective preference features by sampling from the feature distribution, 
effectively capturing both core traits and the subtle, multi-faceted variations within each user group.

\textbf{Preference Conflicts in Unified Modeling.} 
Another major issue is that existing approaches attempt to satisfy multiple preferences with a single output~\citep{yao2024no, park2024rlhf, zhang2024diverging,zhou2024wpo,li2025uncertainty}, typically in the context of textual preferences, such as \textit{``generating responses that are both concise and positive''}.
However, meeting multiple image preferences at once is more challenging.
For example, as shown in Figure~\ref{fig:figure1} (b), conflicting preferences like ``natural scenes" vs. ``sports field" backgrounds are difficult to satisfy simultaneously. 
To move beyond this limitation,
we propose Preference-Conditioned Image Generation (PCIG), which generates tailored images for each group from a unified model, applying a ``one size fits many" approach. 
PCIG first employs a Group-aware Multimodal Large Language Model (G-MLLM) to acquire image generation capabilities and understand group-specific characteristics through targeted pre-training tasks leveraging PAAG-derived user groups. 
Subsequently, we achieve group-wise preference alignment by fine-tuning the G-MLLM with our proposed Group Direct Preference Optimization (Group-DPO), which improves the CTR of generated images for each group.

Beyond these technical challenges, empirical research has been constrained by the absence of large-scale datasets capturing real-world user group-wise image preferences.
To fill this gap, we construct the first large-scale \textbf{G}rouped \textbf{A}dvertising \textbf{I}mage \textbf{P}reference Dataset (\textbf{GAIP}) from a major e-commerce platform, containing approximately 600,000 user groups across 40 million users, with comprehensive group-wise preference annotations.
Our dataset is specifically curated with high image diversity per product and explicit group-wise preference data, substantially exceeding existing datasets in user and group counts, thus providing a robust foundation for developing and evaluating group-aware image generation models.

Our contributions are summarized as follows:

\begin{itemize}

    \item We introduce PAAG, a dynamic user grouping approach that learns product-aware user preferences and derives rich collective features through adaptive clustering and feature sampling, overcoming static grouping limitations in large-scale advertising image generation.

    \item We propose PCIG, a preference-conditioned image generation framework that first enhances a G-MLLM with specialized pre-training tasks and then aligns it with group preferences via Group-DPO. PCIG enables a unified model to generate tailored images for diverse user groups while resolving their preference conflicts.

    \item We present the first-of-its-kind large-scale dataset from a major e-commerce platform, with about 600K groups formed from nearly 40M users and comprehensive group-wise image preference annotations, serving as a foundation for group-wise image generation. Extensive experiments show that our method achieves state-of-the-art results in both online and offline settings.
\end{itemize}

%% file: sec/2_related_work.tex
\section{Related Work}
\label{sec:relatedwork}

\subsection{Advertising Image Generation} 
Advertising image generation focuses on creating customized backgrounds for products to enhance their visual appeal and contextual relevance.
Early approaches for advertising poster and image generation~\citep{Rombach2022HighResolutionIS,zhang2023adding, lu2025uni, whh, lfh, gao2025postermaker,li2024aligning,dzb, fan2025autopp} primarily optimized for offline metrics such as aesthetic appeal, resulting in a gap between visual quality and business impact. 
More recent efforts have shifted attention toward online performance metrics, especially click-through rate (CTR), with pioneering frameworks~\citep{chen2025ctr, shopee} such as CAIG~\cite{chen2025ctr} successfully integrating reinforcement learning to directly maximize CTR. These works demonstrate the potential of learning-based optimization for bridging the gap between visual generation and commercial objectives.  
Nevertheless,  existing approaches still follow a \textit{``one-size-fits-all''} paradigm, where optimization targets aggregated CTR across the entire user population. Such strategies neglect the substantial heterogeneity of preferences among different user groups, which can significantly affect advertising outcomes. To overcome this limitation, our work extends the CTR-driven paradigm by introducing a group-wise RLHF framework that explicitly models group-specific preferences and resolves conflicts, enabling more effective advertising image generation.

\subsection{Reinforcement Learning from Human Feedback}
Reinforcement Learning from Human Feedback (RLHF)~\citep{christiano2017deep} has emerged as a widely used paradigm for aligning generative models with human preferences. 
Classical implementations, such as PPO~\cite{schulman2017proximal}, rely on policy optimization with carefully designed reward models, while more recent variants~\citep{rafailov2023direct,choi2024self} like DPO improve training stability and efficiency by simplifying the optimization pipeline. 
Building on this foundation, personalized RLHF has been proposed to better capture individual tastes and user-specific requirements~\cite{ramesh2024group, chakraborty2024maxmin, li2024personalized, gao2024towards, xiao2024comprehensive, xu2025personalized,poddar2024personalizing}. 
Although such approaches provide fine-grained alignment, fully one-to-one personalization remains computationally expensive and impractical in large-scale real-world deployments~\citep{li2018algorithms}, where models must serve millions of users with diverse behaviors.  
To address this limitation, researchers have turned to group-wise RLHF~\cite{yao2024no, park2024rlhf, zhang2024diverging, liu2025survey, kompatscher2025interactive}, which clusters users into groups and learns group-specific preference models as a compromise between scalability and personalization. 
Despite promising results, current group-wise methods still suffer from two notable issues. 
First, user groupings are often predefined and static, which makes them inflexible when applied to heterogeneous product domains or dynamically shifting user bases. 
Second, inter-group preference conflicts are typically handled through a single aggregated optimization objective, which tends to obscure differences and fails to strike a fair balance among competing preferences.  
Our work directly tackles these challenges by introducing adaptive, product-aware grouping mechanisms together with a preference-aware optimization process, enabling more effective alignment with diverse user groups while preserving scalability. 

%% file: sec/3_method.tex
\section{Method}
\begin{figure*}
    \centering
    \includegraphics[width=0.9\linewidth]{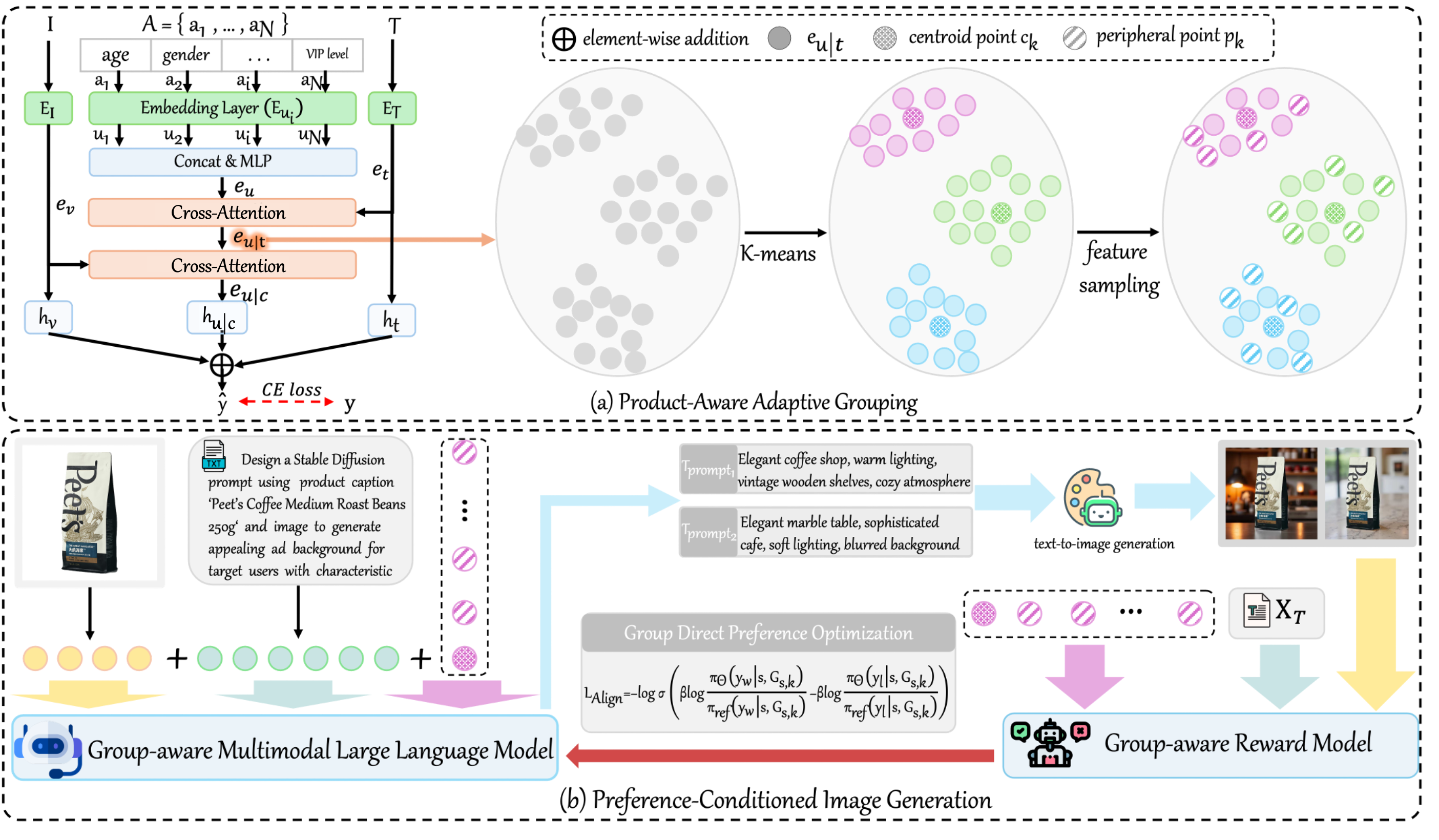}
    \caption{Overview of the OSMF framework: (a) Product-Aware Adaptive Grouping (PAAG) organizes users into preference-coherent groups; (b) Preference-Conditioned Image Generation (PCIG)  tailors advertising images for each group using Group-DPO.}
    \label{fig:overview}
\end{figure*}

\subsection{Overview}

As illustrated in Figure~\ref{fig:overview}, the proposed OSMF framework aligns advertising image generation with diverse group preferences. 
First, Product-Aware Adaptive Grouping (PAAG) forms adaptive, preference-based user groups from product contexts. 
Subsequently, Preference-Conditioned Image Generation (PCIG) leverages these groups to generate tailored content, resolving preference conflicts via group-wise optimization.
In what follows, we elaborate on the design of PAAG and PCIG and how they interact to realize group-wise preference alignment at scale.

\subsection{Product-Aware Adaptive Grouping}
\label{sec-user-clustering}
To address the limited scalability of static user grouping methods, we propose adaptive user clustering based on user attributes and product characteristics. 
As shown in Figure~\ref{fig:overview} (a), PAAG operates through two stages: (1) modeling user preferences for image clicks via multi-modal feature learning and cross-modal interaction, and (2) extracting user image preferences and aggregating them into group features.

\paragraph{\(\rhd\)\;User Preference Modeling.}
To decompose user click behavior into interpretable components, we model distinct preference factors through three key branches: user attributes $A = \{a_1, \dots, a_N\}$ for personalized interaction, advertising image $I$ for visual appeal, and product title $T$ for product affinity.
For image and text modalities, we employ dedicated encoders $E_I$ and $E_T$ to extract semantic features:
\begin{equation}
\mathbf{e}_v = E_I(I) \in \mathbb{R}^d, \quad \mathbf{e}_t = E_T(T) \in \mathbb{R}^d.
\end{equation}
For the user attributes, we first obtain the dense vector $\mathbf{u}_i \in \mathbb{R}^{d'}$ by mapping each attribute $a_i \in A$ via a unique embedding layer. $\mathbf{u}_i$ are then concatenated and pass through a Multi-Layer Perceptron (MLP) to capture the complex inter-dependencies, yielding the  user embedding $\mathbf{e}_u \in \mathbb{R}^d$.


To capture dynamic user preferences that adapt to specific product contexts, we first create a product-aware user representation by allowing user features to attend to textual product information:
\begin{equation}
\mathbf{e}_{u|t} = \mathbf{CA}(Q=\mathbf{e}_u, K=\mathbf{e}_t, V=\mathbf{e}_t),
\label{eq:e_u_t}
\end{equation}
where \textit{CA} refers to the cross-attention layer, and $\mathbf{e}_{u|t}$ could encode user preferences conditioned on the specific product being evaluated. 
Subsequently, we capture visual preferences by letting the product-conditioned user features attend to image representations:
\begin{equation}
\mathbf{e}_{u|c}  = \mathbf{CA}(Q=\mathbf{e}_{u|t}, K=\mathbf{e}_v, V=\mathbf{e}_v).
\label{eq:e_u_c}
\end{equation}
The final embedding $\mathbf{e}_{u|c}$ encapsulates user click preferences that are jointly conditioned on both product semantics and visual characteristics.

Based on the learned embeddings, each branch employs a dedicated prediction head $h(\cdot)$ to generate scalar logits:
\begin{equation}
\hat{y} = h_t(\mathbf{e}_t) + h_v(\mathbf{e}_v) + h_{u|c}(\mathbf{e}_{u|c} ).
\end{equation}
The model is trained end-to-end by minimizing the cross-entropy loss between predicted clicks $\hat{y}$ and ground-truth $y$.

\paragraph{\(\rhd\)\;Group Feature Aggregation.} 
\label{par-user-clustering}
Building on the trained preference model, we then cluster users to capture similar preference patterns for each product. 
For a given product $s$, we first extract the product-aware representations $\mathcal{E}_s = \{\mathbf{e}_{u|t} \mid u \in \mathcal{U}_s\}$ for all users $\mathcal{U}_s$ who have previously viewed product $s$.
Next, we apply K-Means clustering on $\mathcal{E}_s$ to find the partition $C = \{C_1, \ldots, C_K\}$ that minimizes the within-cluster sum of squares for that product's user set:
\begin{equation}
\label{eq:kmeans_product}
    \underset{C}{\operatorname{argmin}} \sum_{k=1}^{K} \sum_{\mathbf{e}_{u|t} \in C_k} \|\mathbf{e}_{u|t} - \mathbf{\mu}_k\|^2,
\end{equation}
where $\mathbf{\mu}_k$ is the centroid of cluster $k$. The optimal number of clusters $K_s^*$ for product $s$ is determined by maximizing the mean silhouette coefficient for different values of $K$.

For each resulting group $C_k$ of product $s$, we construct a rich collective representation $G_{s,k}$ that captures intra-group diversity. Due to the natural variations in user preferences within clusters, a single centroid may inadequately represent the full preference spectrum. Hence, we select multiple representative points at varying distances from $\mathbf{\mu}_k$ to form a comprehensive group embedding: 
\begin{equation}
\mathbf{G}_{s,k} = \{\mathbf{\mu}_k\} \cup \{\mathbf{p}_{k}^{(j)} \mid j \in \mathcal{J}\}.
\end{equation}
Here, $\mathbf{p}_{k}^{(j)}$ denotes points sampled at percentile distances indexed by $\mathcal{J}$, with denser sampling in outer regions.
This ensures that both the core preference (represented by the centroid) and the surrounding preference  (peripheral points $\mathbf{p}_{k}^{(j)}$) are effectively incorporated, enabling more nuanced group-based preference modeling in subsequent stages. 


\subsection{Preference-Conditioned Image Generation}
\label{sec-capm}
Based on the rich collective features derived from PAAG, we propose the \textbf{G}roup-aware \textbf{M}ultimodal \textbf{L}arge \textbf{L}anguage \textbf{M}odel (G-MLLM), which integrates group-level preferences into a single MLLM through two tailored training objectives, and further enhances alignment via Group-wise Preference Alignment. This enables the generation of more personalized visual content beyond the traditional \textit{``one-size-fits-all''} paradigm.

\paragraph{\(\rhd\)\;Group-Aware Multimodal Large Language Model.}
To enable G-MLLM to perceive group preferences, we integrate the group information into the model.
Specifically, we achieve this by
prepending a group token $\mathbf{e}_{G_{s,k}} $ for group $k$ to the standard MLLM input sequence, which is obtained by encoding $G_{s,k}$ by an MLP. In a typical MLLM, the input $\mathbf{X}_{\text{input}}$ consists of visual tokens $\mathbf{X}_V$ and textual tokens $\mathbf{X}_T$. Our approach enhances this by incorporating group-specific preferences:
\begin{equation}
    \mathbf{X}_{\text{input}} = [\mathbf{e}_{G_{s,k}}; \mathbf{X}_V; \mathbf{X}_T].
\end{equation}

Substantially, we design a series of group-centric tasks including group analysis and behavioral prediction to improve G-MLLM's understanding of group characteristics. Meanwhile, the product-centric tasks including the product comprehension and prompt generation are further adopted to promote the capacity of generating high-quality prompts for advertising images. The targets of each kind of tasks are illustrated as follows:

\begin{itemize}
    \item \textit{Group analysis} aims to help G-MLLM characterize the typical user feature within $C_k$ through describing the most common user profile in natural language. 
    \item \textit{Behavioral prediction} intends to encourage G-MLLM to model the fine-grained similarity within users clicking the same product advertisement by predicting the next possible clickers based on the existing ones.
    \item \textit{Product comprehension} contains predicting the product title based on the advertising image or product image, by which G-MLLM learns to understand the image in the e-commerce style better.
    \item \textit{Prompt generation} involves generating image prompt tailored for specific users based on the product image or title. The G-MLLM is substantially anticipated to generate appropriate advertising images adaptively.
\end{itemize}
Please refer to the supplementary materials for more details on the pre-training tasks and the instruction pool.

\paragraph{\(\rhd\)\;Group-wise Preference Alignment.}
After pre-training, we align the G-MLLM with real-world user engagement through group-aware preference optimization. To establish preference pairs of advertising images for optimization, two key issues remain to be addressed: rendering the textual prompt generated by G-MLLM into a high-fidelity advertising image, and evaluating the group-wise click preference of any pairs of generated images during training. 

In this work, we combine ControlNet-Inpaint~\cite{zhang2023adding} and Stable Diffusion~\cite{Rombach2022HighResolutionIS} as the text-to-image model to generate advertising images for products (the “\textit{text-to-image generation}” part shown in Figure~\ref{fig:overview}). Subsequently, we extend G-MLLM to the Group-aware Reward Model (GRM), so as to evaluate visual quality with awareness of group-specific preferences. 
To train the GRM, we construct pair-wise samples from the proposed GAIP dataset,
where each pair consists of two advertising images for the same product with CTR rankings. Specifically, GRM takes image pair, group embedding and corresponding instructions $X_T$ as input, and the preference is expected to be encoded into the hidden states of G-MLLM $H_{\text{GRM}}$. Then, following the binary classification approach, we employ a classification head $FC_{\text{cls}}$ to predict the relative CTR performance:
\begin{equation}
\mathbb{P}(\text{CTR}_1 > \text{CTR}_2|G_{s,k}) = \text{Softmax}(FC_{\text{cls}}(H_{\text{GRM}})),
\end{equation}
where $\text{CTR}_n$ denotes the CTR of the $n$-th image when exposed to  group  $C_k$. This group-aware design enables the reward model to provide more accurate preference signals tailored to specific user groups, which is crucial for effective preference alignment in our framework.

Using the GRM predictions, we construct group-specific preference tuples $(s, G_{s,k}, y_w, y_l)$ for each context (product $s$, group $C_k$), where $y_w$ and $y_l$ are prompts corresponding to winner and loser images $I_w$ and $I_l$ based on their CTR performance within $C_k$. The G-MLLM is then fine-tuned using Group-DPO to maximize preference for winning prompts:

\begin{equation}
\label{eq:align_loss}
\begin{split}
\mathcal{L}_{\text{Align}} = - \log \sigma\bigg( & \beta \log \frac{\pi_{\theta}(y_w|s, G_{s,k})}{\pi_{\text{ref}}(y_w|s, G_{s,k})} \\
& - \beta \log \frac{\pi_{\theta}(y_l|s, G_{s,k})}{\pi_{\text{ref}}(y_l|s, G_{s,k})}\bigg),
\end{split}
\end{equation}
where $\sigma(\cdot)$ is the sigmoid function, $\pi_{\theta}(y|s, G_{s,k})$ models the probability of generating prompt $y$ conditioned on both product $s$ and group $G_{s,k}$, $\pi_{\text{ref}}$ is the reference pre-trained G-MLLM, and $\beta$ is the temperature parameter. This process directly optimizes the G-MLLM to generate prompts that yield higher CTR for specific target groups, effectively resolving cross-group preference conflicts.

%% file: sec/4_dataset.tex
\section{Dataset}
Empirical research on group-wise advertising preferences is fundamentally constrained by the lack of large-scale, real-world datasets. 
Existing public datasets are either limited in scale~\citep{ramesh2024group, jang2023personalized,li2024aligning} or lack essential user-creative interaction signals~\cite{shopee, chen2025ctr}, rendering them insufficient for modeling real-world group behaviors.
To bridge this gap, we introduce the Grouped Advertising Image Preference (GAIP) dataset, the first large-scale, meticulously annotated dataset tailored for group-wise image preference modeling, constructed from a major e-commerce platform and designed to enable scalable analysis of preference heterogeneity across user groups.

\subsection{Dataset Construction}

GAIP originates from the industrial advertising logs of the e-commerce platform. Firstly, we elaborately record the product information (advertising image and title), the user profile (e.g., purchasing power, age, promotion sensitivity) and behavior (clicking the advertisement or not) in a three-week period. The resulting user-advertisement interactions dataset covers 40,027,535 users, 2,085,969 products and 9,565,154 advertising images, with high image diversity per product.
Subsequently, users are divided into $k$ distinct groups for each product based on PAAG, and we calculate CTR of each advertising image for each user group. Then we develop GAIP by constructing quadruples $(I^{i}_{s}, T_{s}, G_{{s},k}, CTR(I_s^i,k))$, where $I_{s}^i$ and $T_s$ are the $i_{th}$ advertising image and title of product $s$. $CTR(I_s^i,k)$ represents the CTR of $I_{s}^i$ in the $k_{th}$ group, whose group embedding is $G_{{s},k}$. 
To ensure statistical confidence, we filter low-exposure products, yielding 610,172 user groups.

\subsection{Dataset Characteristics}

Compared to existing publicly available datasets, the dataset we propose offers two key advantages:

\begin{itemize}[leftmargin=*, labelindent=5pt]
    \item \noindent \textit{Fine-Grained:} Unlike other CTR-driven image generation datasets~\cite{shopee, chen2025ctr} that provide aggregated or broad user preferences, GAIP offers fine-grained, group-level preference data. Its granularity enables a deeper understanding of inter-group differences in click preferences, facilitating the generation of highly targeted advertising images tailored to different user groups.
    
    \item \noindent \textit{Large-Scale:} Most publicly available datasets on group-wise preference~\cite{chakraborty2024maxmin, ramesh2024group} are limited in scale (covering less than 10 groups). In contrast, GAIP contains around 600,000 real-world group-wise preference data, which supports more challenging, reliable and practically valuable research on group-wise preference.

\end{itemize}
Please see the supplementary materials for more details.

%% file: sec/5_experiment.tex
\section{Experiment}

\subsection{Implementation Details}
In PAAG, image and text features are extracted via ResNet~\citep{he2016deep} and CLIP's text encoder~\cite{radford2021learning}, respectively, and then mapped to 128-dimensional embeddings using learnable linear layers.
To ensure computational efficiency, the number of user groups clustered per product is limited to a maximum of 5. 
Group features are aggregated via a percentile-based sampling strategy, selecting values at the 
15th (twice), 55th (three times), and 95th (five times) percentiles for broad distribution coverage.
For PCIG, we adopt LLaVA~\citep{liu2023visual} as the backbone. 
During G-MLLM pre-training on our task suite, we train the full model for 10 epochs with a cosine schedule (LR $=$ 2e-6), which takes about ~5 days.
To train GRM, we first reconstruct GAIP to obtain pair-wise advertising images of the same product. Then we initialize GRM with G-MLLM and train it with the same setup. 
In the final Group-DPO stage, we freeze GRM and apply LoRA~\citep{hu2022lora} to fine-tune G-MLLM for 3 epochs with a LR of 2e-5, taking ~50 hours. 
All experiments run on a single 8×NVIDIA H100 node.

\subsection{Evaluation of Group Aggregation Effects}

\paragraph{\(\rhd\)\;Baselines and metrics.}
To validate the proposed PAAG, we compare against three recent user preference modeling methods: CACS~\citep{lin2022joint}, WIYD~\citep{you2023image}, and JAC~\citep{yang2024parallel}.
For evaluation, we employ two standard metrics. First, to assess group aggregation effectiveness, we utilize NDCG@5~\citep{jarvelin2002cumulated}, which quantifies the divergence in CTR-based ranking patterns across different user groups. 
For each product, we compute NDCG@5 between every pair of groups over a shared set of candidate advertising images, treating one group’s ranking as reference and the other’s as prediction, and then average over all products.
The \textit{“@5”} denotes that the top-5 ranked creatives are used in the computation. In our setting, lower NDCG@5 indicates more distinct group preferences.
Second, we adopt AUROC~\citep{hanley1982meaning} as an auxiliary metric to evaluate the quality of user preference modeling. Higher AUROC reflects better distinction between clicked and non-clicked items.
All metrics are computed on clustering results from 1{,}000 representative products, covering about 100{,}000 samples.

\begin{table}[t]
\centering
\resizebox{0.65\linewidth}{!}{
\begin{tabular}{lcc}
\toprule
\textbf{Model} & \textbf{NDCG@5↓} & \textbf{AUROC↑} \\
\midrule
CACS~\citep{lin2022joint} & \underline{0.3124} & 0.6188 \\
WIYD~\citep{you2023image} & 0.3165 & \underline{0.6213} \\
JAC~\citep{yang2024parallel} & 0.3193 & 0.6179 \\
\textbf{Ours} & \textbf{0.3066} & \textbf{0.6372} \\
\bottomrule
\end{tabular}
}
\caption{Performance comparison on preference modeling against state-of-the-art baselines. Bold and underlined denote the best and second-best results, respectively.}
\label{tab:sota_comparison}
\end{table}

\paragraph{\(\rhd\)\;Main Results.}
As shown in Table~\ref{tab:sota_comparison}, our method achieves superior performance on both metrics.
Concretely, PAAG attains the lowest NDCG@5 (0.3066), outperforming the best baseline (CACS) 
, indicating more distinct inter-group preference patterns for effective group-wise advertising generation. 
Additionally, PAAG achieves the highest AUROC (0.6372), improving over the strongest baseline (WIYD) by 0.0159. 
This validates that our user preference modeling provides more reliable preference estimation, forming the foundation for group aggregation. The dual improvements demonstrate that enhanced group-wise preference modeling directly benefits group aggregation.

\paragraph{\(\rhd\)\;Ablation Study.}
As shown in Table~\ref{tab:ablation_study}, we conduct a comprehensive ablation study to validate the contribution of each component in our framework. The progressive addition reveals clear insights:

\begin{table}[t]
\centering
\resizebox{0.9\columnwidth}{!}{
\begin{tabular}{ccccc|cc}
\toprule
\textbf{I} & \textbf{U} & \textbf{T} & \textbf{UT} & \textbf{UI} & \textbf{NDCG@5↓} & \textbf{AUROC↑} \\
\midrule
\ding{51} & \ding{55} & \ding{55} & \ding{55} & \ding{55} & -- & 0.5974 \\
\ding{51} & \ding{51} & \ding{55} & \ding{55} & \ding{55} & 0.3211 & 0.6098 \textit{(+0.0124)} \\
\ding{51} & \ding{51} & \ding{51} & \ding{55} & \ding{55} & 0.3193 \textit{(-0.0018)} & 0.6104 \textit{(+0.0130)} \\
\ding{51} & \ding{51} & \ding{51} & \ding{51} & \ding{55} & 0.3084 \textit{(-0.0127)} & 0.6207 \textit{(+0.0233)} \\
\ding{51} & \ding{51} & \ding{51} & \ding{51} & \ding{51} & 0.3066 \textit{(-0.0145)} & 0.6372 \textit{(+0.0398)} \\
\bottomrule
\end{tabular}
}
\caption{Ablation study of PAAG. \textbf{I}: image-level CTR; \textbf{U}: user-level CTR; \textbf{T}: title-level CTR; \textbf{UT}: product-aware user-text interaction; \textbf{UI}: product-aware user-image interaction. ``--'' indicates NDCG@5 unavailable due to lack of user info.}
\label{tab:ablation_study}
\end{table}

\begin{itemize}[leftmargin=*, labelindent=5pt]
   
    \item \textit{Progressive Factorization of CTR:} We progressively add independent branches to factorize CTR, as shown in Figure~\ref{fig:overview}(a).
    Starting with $e_v$ (AUROC: 0.5974), adding $e_u$ improves AUROC to 0.6098 (+0.0124) and NDCG@5 to 0.3211. Further adding $e_t$ achieves AUROC of 0.6104 (+0.0130) and NDCG@5 of 0.3193. This demonstrates each component's incremental contribution to click preference prediction and group feature aggregation. 

    \item \textit{User-Text Interaction:} Adding the user-title interaction branch ($e_u$ and $e_t$ interaction as shown in Eq.~\ref{eq:e_u_t}) significantly improves both AUROC to 0.6207 and NDCG@5 to 0.3084. This indicates that capturing interactions between user preferences and product semantics enhances both click prediction accuracy and group distinctiveness.

    \item \textit{User-Visual Interaction:} Adding the user-image interaction branch ($e_u$ and $e_v$ interaction as shown in Eq.~\ref{eq:e_u_c}) completes the full PAAG model, achieving AUROC of 0.6372 and NDCG@5 of 0.3066. This confirms that integrating visual information into user representations enhances click preference prediction and user grouping.
\end{itemize}

\begin{figure}[t]
    \centering
    \includegraphics[width=0.94\columnwidth]
    {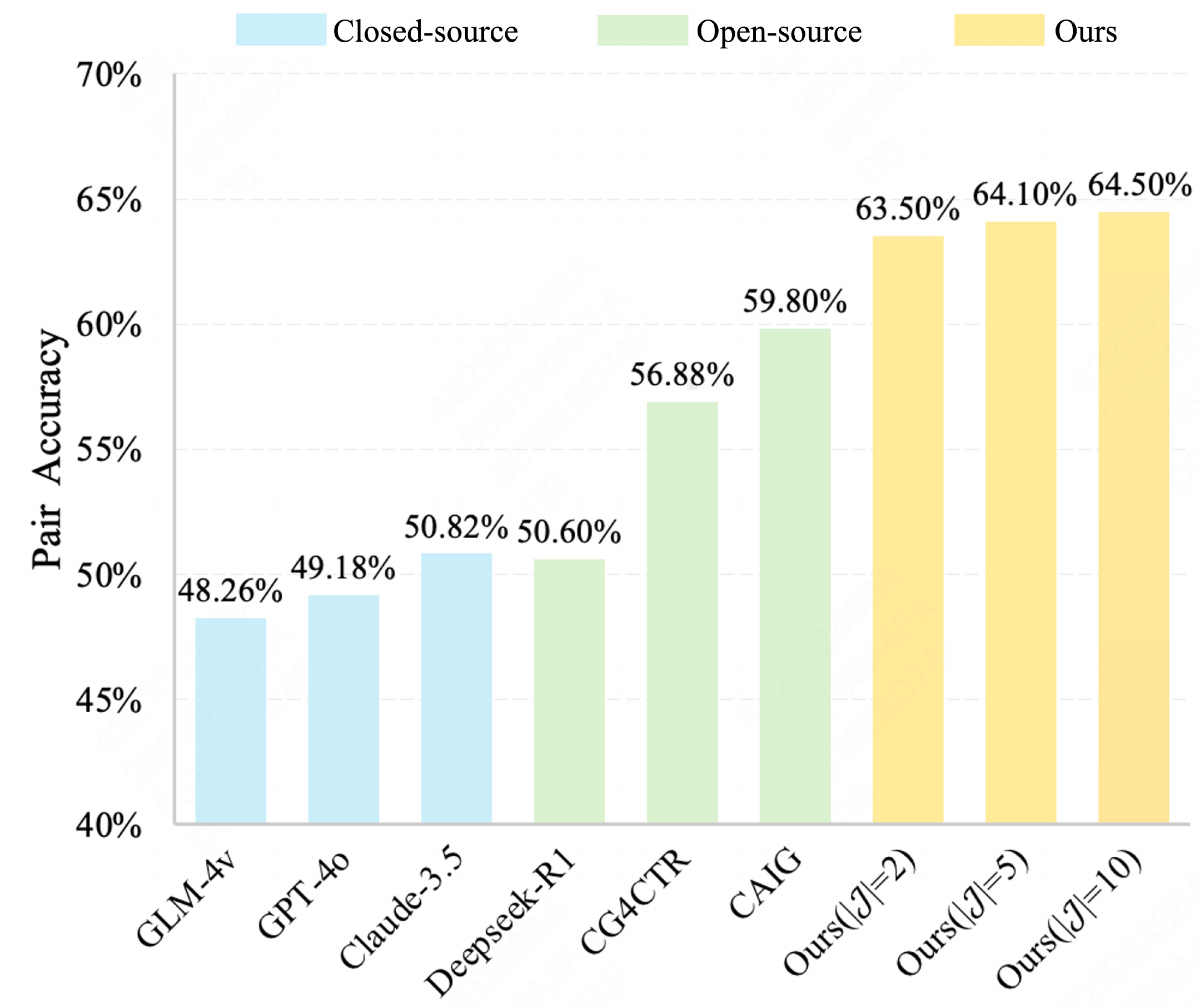}
    \caption{Comparison with SOTA reward models.}
    \label{fig:pair_accuracy}
\end{figure}

\begin{figure*}[t]
    \centering
    \includegraphics[width=1\textwidth]
    {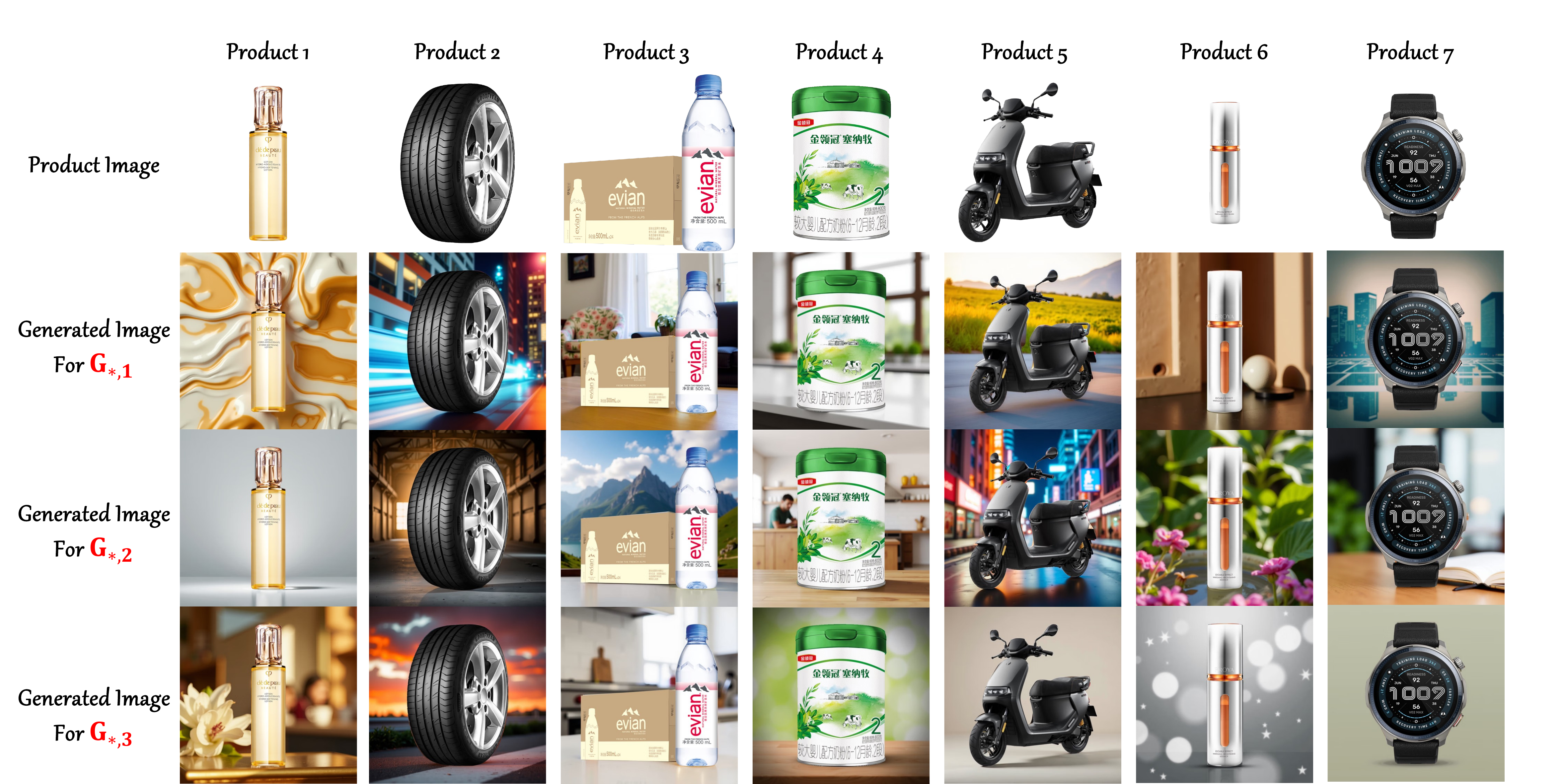}
    \caption{Qualitative results of OSMF. $\mathbf{G}_{*,k}$ denotes the $k$-th group feature for product ``$*$''. Group assignments are product-dependent, so rows correspond to different groups.}
    \label{fig:qualitative-results}
\end{figure*}

\subsection{Evaluation of Group Alignment Effects}
\paragraph{\(\rhd\)\;Baselines and metrics.}
We evaluate group alignment from two perspectives: online CTR for PCIG and offline reward modeling for GRM. 
For online CTR, we compare PCIG with three CTR-driven baselines: the pretrained G-MLLM, CG4CTR~\cite{shopee}, and CAIG~\cite{chen2025ctr}. 
For each product $s$ with $K^*_s$ user groups, we generate one customized image per group and expose it only to its target group, whereas group-agnostic methods use $K^*_s$ images shown to all groups. 
We evaluate CTR over a one-week period on 10{,}000 high-exposure products, reporting both absolute CTR and relative gains over the pretrained G-MLLM.
For offline reward modeling, we evaluate GRM with Pair Accuracy~\cite{chen2025ctr}, which measures how well a model ranks CTR between image pairs conditioned on the same group embedding. 
We compare GRM against two types of baselines: CTR reward models from CAIG~\cite{chen2025ctr} and CG4CTR~\cite{shopee}, and strong general-purpose MLLMs, including GPT-4o~\citep{achiam2023gpt}, Claude 3.5 Sonnet~\citep{claude2024}, GLM-4v~\citep{glm2024chatglm}, and DeepSeek R1~\citep{guo2025deepseek}\footnote{Since DeepSeek does not support image inputs, we use GPT-4o to convert each image into text descriptions before comparison.}. 
We evaluate on over 1{,}200 image pairs with their target user group.

\paragraph{\(\rhd\)\;Main results.} Table~\ref{tab:ctr} reports the online CTRs of PCIG and CTR-driven counterparts~\cite{chen2025ctr, shopee}. Since absolute CTR values in recommendation settings are often low, we report both the raw CTR and the relative improvement over the pretrained G-MLLM baseline. Both CAIG and PCIG significantly improve CTR (by approximately 3.4\% and 5.5\%, respectively), demonstrating the effectiveness of preference modeling. Notably, PCIG also surpasses CAIG by a clear margin. CAIG optimizes for the aggregate user base,  failing to capture group-specific preferences, whereas Group-DPO in PCIG explicitly models group-level preferences, enabling highly customized image generation that resonates with each user group. In practice, even marginal CTR improvements (e.g., 0.5\%) can translate into substantial revenue gains, underscoring the practical value of our approach. For reward modeling, Figure~\ref{fig:pair_accuracy} compares GRM with the above CTR reward baselines and general-purpose MLLMs. These models exhibit limited performance in selecting the higher-CTR image from input pairs, highlighting the domain-specific nature of CTR prediction and the difficulty of evaluation. In contrast, GRM significantly outperforms all counterparts, achieving a 4.7\% improvement in Pair Accuracy over CAIG. This gain largely stems from the incorporation of user group information, which provides crucial context for preference estimation. Since reward quality is central to Group-DPO, GRM’s improved predictive accuracy is a key driver of PCIG’s overall performance gains.
\begin{table}[t]
\centering
\small
\begin{tabular}{lcc}
\toprule
\textbf{Method} &\textbf{CTR} & \textbf{Improvement} \\
\midrule
Pretrained G-MLLM & 0.0146 & - \\
CG4CTR~\cite{shopee} & 0.0143 & -2.055\% \\
CAIG~\cite{chen2025ctr} & \underline{0.0151} & +3.425\% \\
Ours (w/o group) & 0.0148 & +1.370\% \\
\textbf{Ours} & \textbf{0.0154} & \textbf{+5.479\%} \\
\bottomrule
\end{tabular}
\caption{Online CTR results of different methods.}
\label{tab:ctr}
\end{table}

\paragraph{\(\rhd\)\;Ablation Study.}
We conduct two ablation studies to examine (1) group feature modeling in G-MLLM and (2) group feature aggregation in PAAG/GRM, with results summarized in Table~\ref{tab:ctr} and Figure~\ref{fig:pair_accuracy}, respectively. We detail these ablations below:

\begin{itemize}[leftmargin=*, labelindent=5pt]
    \item \textit{Effect of group feature modeling in G-MLLM:}
    We replace G-MLLM with a vanilla MLLM by removing the group feature $\mathbf{e}_{G_{s,k}}$ from the input sequence. This variant is pretrained on product-centric tasks and fine-tuned with Group-DPO, denoted as \textit{ours w/o group}. As shown in Table~\ref{tab:ctr}, although Group-DPO improves CTR by about 1.37\% over the pretrained model, its performance still falls well behind the full G-MLLM and CAIG. This highlights the importance of explicitly modeling group-specific preferences via $\mathbf{e}_{G_{s,k}}$ to distinguish and adapt to diverse, sometimes conflicting, group interests.
    \item \textit{Effect of group feature aggregation in PAAG/GRM:}
    We vary the size of $\mathcal{J}$ during group feature extraction. Without $\mathcal{J}$ and $\mathbf{\mu}_k$, GRM degenerates to the reward model in CAIG. 
    As shown in Figure~\ref{fig:pair_accuracy}, increasing the number of sampled users consistently improves GRM performance, as richer group features capture more detailed group signals.
    However, an excessively large $\mathbf{\mu}_k$ incurs substantial storage and computation costs, so we set the capacity of $\mathcal{J}$ to 10 as a trade-off between effectiveness and efficiency.
\end{itemize}

\paragraph{\(\rhd\)\;Qualitative Results.}
We illustrate the qualitative results in Figure~\ref{fig:qualitative-results}, to show the diverse but attractive generated images for different user groups with rich visual styles. 
Taking the cosmetics in the 1st column as an example, PCIG generates images characterized by either vibrant color schemes, minimalist palettes, or photography-inspired visual styles, tailored for three prototypical target user groups. 
Similarly, for other products, PCIG ensures the stylistically diverse images to accommodate the click preferences of distinct user groups, thereby demonstrating its strong capability to adapt generation to heterogeneous user demands and capture subtle, fine-grained preference differences across varied user groups, highlighting its potential for group-aware advertising image generation at scale.

%% file: sec/6_conclusion.tex
\section{Conclusion}

In this paper, we introduce \textit{One Size, Many Fits} (OSMF), a unified framework for advertising image generation that aligns diverse group-wise click preferences, overcoming the limitations of existing \textit{one-size-fits-all} methods. 
OSMF first clusters user features by PAAG, and encodes the group preference into group features. Afterwards, PCIG generates tailored images for each user group. Concretely, we develop the G-MLLM to generate group-wise image prompts and GRM to simulate group preferences. 
On this basis, we present Group-DPO to improve group-level preference alignment in G-MLLM. OSMF consistently outperforms the SOTA counterparts in both online and offline experiments, demonstrating the effectiveness of group-aware image generation.
In addition, we will release GAIP, the first-of-its-kind large-scale dataset for group-wise image preference modeling, containing around 600,000 group preference data, to support future research.
Our work highlights the value of group-wise preferences in improving preference modeling at scale and enabling broader applications.

\section{Acknowledgments}
This work was supported by the National Natural Science Foundation of China (62276256, U2441251), and the Young Elite Scientists Sponsorship Program by CAST (2023QNRC001).

%% file: sec/7_suppl.tex
\clearpage
\maketitlesupplementary

In the supplementary materials, we provide additional resources to support research reproducibility. 
As outlined in the contents above, we include extended visualization results of the Group-aware Multimodal Large Language Model (G-MLLM) framework and dataset characterization (Section A.1), comprehensive descriptions of user profile attributes (Section A.2), the design principles for instruction sets (Section A.3), and detailed algorithmic specifications of the group feature aggregation methodology (Section A.4). 
Additionally, we further discuss current methodological limitations and future research directions (Section A.5), analyze potential social impacts with implemented safeguards (Section A.6), and outline ethical and legal considerations regarding privacy and data usage (Section A.7).
Together, these materials support independent verification and promote rigorous, reproducible research.

\subsection*{A.1 More Visualization Results}
\addcontentsline{toc}{subsection}{A.1 More Visualization Results}
In this section, we provide additional visualization results from three perspectives. First, we demonstrate our method's performance across diverse product categories (Section A.1.1). Second, we show how our approach generates varied backgrounds for the same product tailored to different user groups (Section A.1.2). Finally, we present an overview of our GAIP dataset, showcasing its comprehensive coverage and diversity (Section A.1.3).

\subsubsection*{A.1.1 Visualization of More Products Results }
\addcontentsline{toc}{subsubsection}{A.1.1 Visualization of More Products Results}

\begin{figure*}[h]
   \centering
\includegraphics[width=\textwidth]{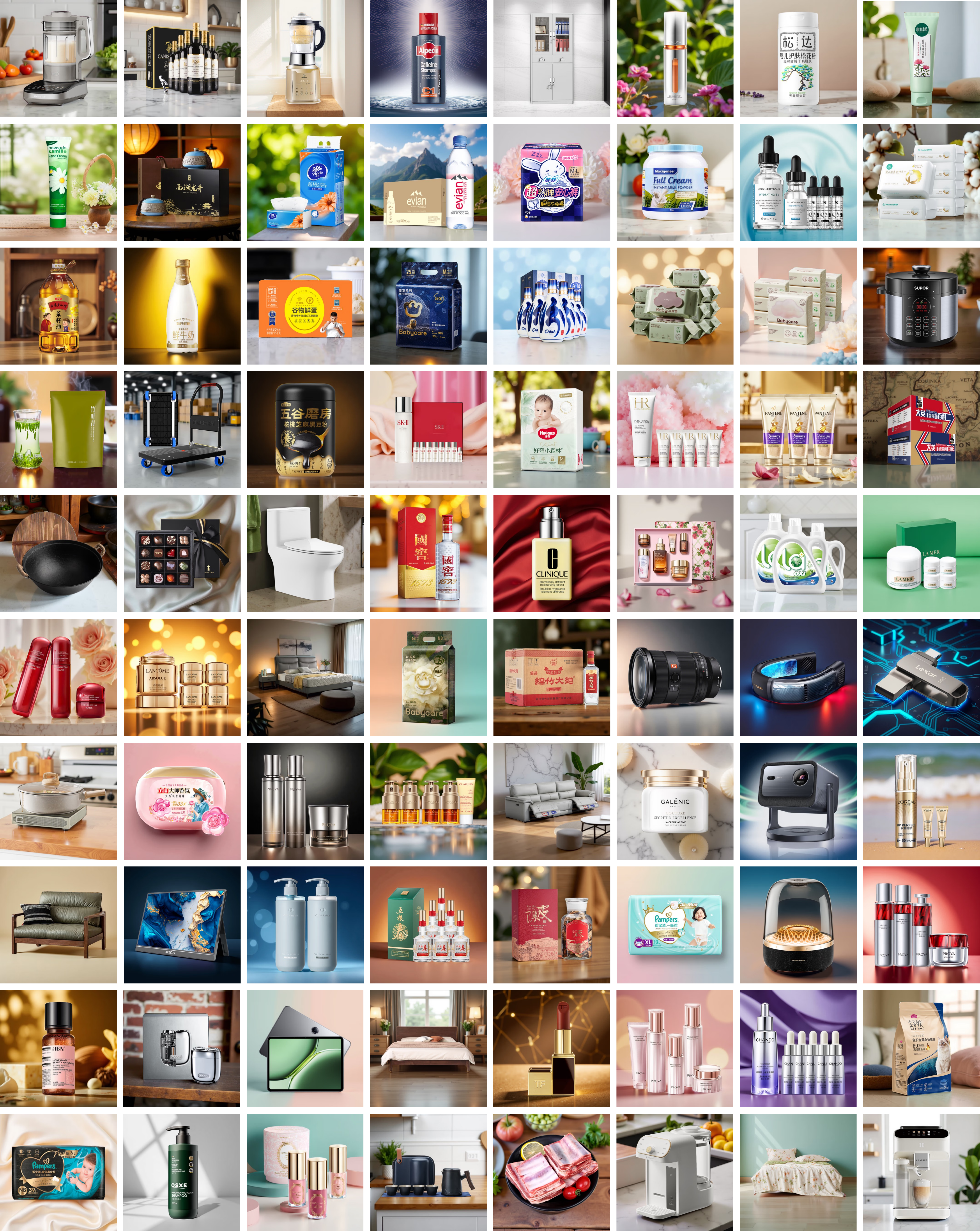}
   \caption{More visualization examples generated by our method across diverse product categories.}
   \label{fig:diff_sku}
\end{figure*}

Figure~\ref{fig:diff_sku} displays a comprehensive set of advertising images generated by our method across a diverse array of products, including electronics, beauty products, daily necessities, and other commonly advertised product types. 
These examples demonstrate the model’s ability to generate contextually relevant and visually compelling backgrounds tailored to the specific characteristics of each product type and the corresponding user group. 
By showing the visualization results, we highlight the effectiveness of our method in accommodating the click preferences of different user groups.

\subsubsection*{A.1.2 Visualization of the Same Products Results}
\addcontentsline{toc}{subsubsection}{A.1.2 Visualization of the Same Products Results}
Figure~\ref{fig:same_sku} presents multiple background generation results for the same product. This visualization illustrates our method’s ability to create diverse and contextually relevant backgrounds tailored to different user groups. The generated backgrounds reflect variations in group-wise preferences, underscoring the flexibility of our method in adapting to distinct group-wise features while maintaining visual coherence and product relevance.

\subsubsection*{A.1.3 Visualization of Dataset}
\begin{figure*}[h]
   \centering
\includegraphics[width=\textwidth]{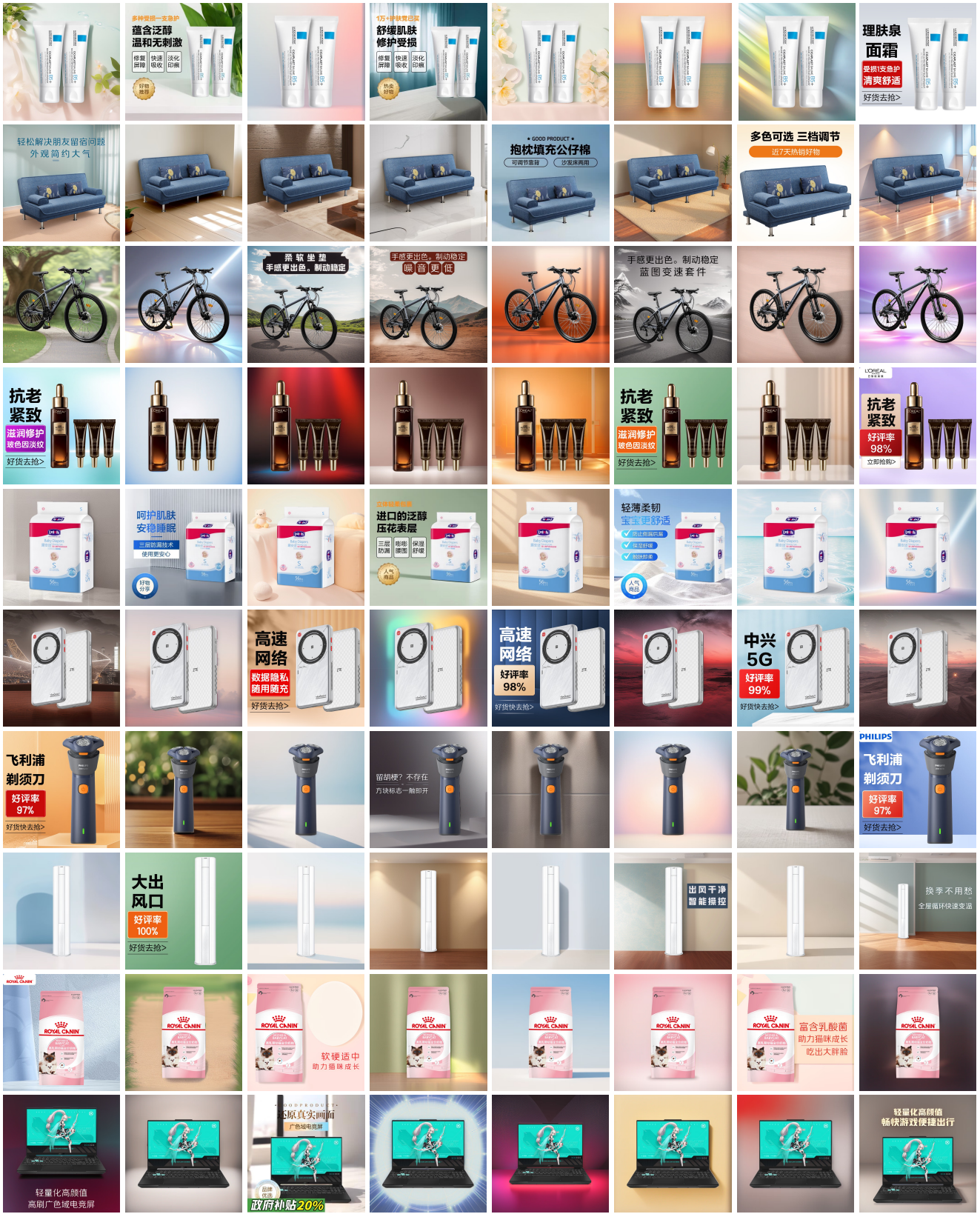}
   \caption{More visualization examples from our dataset spanning a wide range of product categories, with each row illustrating diverse creative presentations for the same product.}
   \label{fig:datset}
\end{figure*}

\addcontentsline{toc}{subsubsection}{A.1.3 Visualization of Dataset}
Figure~\ref{fig:datset} provides an overview of the Grouped Advertising Image Preference dataset (GAIP), illustrating representative samples across a broad spectrum of product categories. The visualization highlights the dataset's comprehensive coverage and diversity, which are essential for capturing a wide range of product contexts and user scenarios.
Built from real-world user interaction data collected from a major e-commerce platform, GAIP enables scalable analysis of heterogeneous user preferences across diverse product categories. We will release GAIP to foster future research on group-level preference modeling in advertising and beyond.

\begin{table}[h!]
\resizebox{\columnwidth}{!}{%
\begin{tabular}{c>{\centering\arraybackslash}p{0.8\linewidth}}
\hline
\textbf{Attribute Name} & \textbf{Description} \\ \hline

age & \RaggedRight User's age group, divided into 13 segments. \\ 

color & \RaggedRight User's preferred product color, with 15 options. \\ 

is\_vip & \RaggedRight Indicates whether the user is a VIP member. \\

os\_plant & \RaggedRight Most frequently used mobile operating system. \\ 

has\_car & \RaggedRight Whether the user owns a car. \\

gender & \RaggedRight User's gender. \\ 

marriage & \RaggedRight User's marital status. \\

child\_age & \RaggedRight Age range of the user's children, divided into 10 segments.\\

has\_house & \RaggedRight Whether the user owns a house.  \\ 

profession & \RaggedRight User's profession, categorized into 11 types.\\ 

user\_active & \RaggedRight User's activity level on the platform (3 levels). \\ 

education & \RaggedRight User's education level (6 categories). \\ 

target\_group & \RaggedRight User's social group, e.g., student, urban family. \\ 

member\_level & \RaggedRight User's membership tier, higher values represent higher tiers. \\ 

customer\_points & \RaggedRight User engagement, categorized into five levels. \\

purchase\_power & \RaggedRight User's purchasing ability, lower values indicate higher power. \\ 

resident\_city\_level & \RaggedRight User's city tier, from Tier 1 to Tier 5. \\ 

per\_month\_order\_num & \RaggedRight User's average monthly paid orders over the last six months.  \\ 

is\_impulse\_purchase\_user & \RaggedRight Whether the user tends to make impulse purchases. \\ 

promotions\_sensitive\_level & \RaggedRight User's sensitivity to promotions; higher values indicate greater sensitivity. \\
\hline
\end{tabular}%
}
\caption{Descriptions of User Profile Attributes.}
\label{user_attri}
\end{table}

\subsection*{A.2 User Profile Attributes}
\addcontentsline{toc}{subsection}{A.2 User Profile Attributes}

In this section, we provide detailed descriptions of the comprehensive user profile attributes employed in our method. As outlined in Table~\ref{user_attri}, these attributes encompass diverse demographic, behavioral, and preference characteristics that enable our G-MLLM to generate personalized advertising images tailored to specific user groups. 
By leveraging this rich set of user features, our approach effectively captures the nuanced preferences and characteristics of different audience segments, resulting in more relevant and engaging advertisement creatives.
It is important to note that these attributes are designed to remain anonymized and do not include any personally identifiable information (PII). They are aggregated and generalized in a way that ensures individual users cannot be specifically identified or located, preserving user privacy and complying with data protection standards.

\subsection*{A.3 Details of  the Instruction Pool}
\addcontentsline{toc}{subsection}{A.3 Details of  the Instruction Pool}

In this section, we present the comprehensive instruction pool constructed for G-MLLM, as detailed in Table~\ref{tab1}. The instruction pool strategically integrates user demographic and behavioral attributes to capture nuanced preference patterns across distinct user segments. We organize the instructions into two primary categories: group-centric tasks and product-centric tasks.

The group-centric tasks include specialized instructions for group analysis and behavioral prediction. For group analysis, we sample real-world user clusters after the product-aware adaptive grouping, and utilize the description of the most common attribute characteristics of users within the cluster as the ground truth. For behavioral prediction, we first collect the clicking users of each advertising image from advertising logs of the e-commerce platform. Then we encourage the G-MLLM to predict the profile of one of these users based on the other users. The product-centric tasks focus on multimodal comprehension and creative prompt generation. Concretely, we firstly sample the products and collect their product images and titles, then make G-MLLM predict the advertising title based on the product image in multimodal comprehension subtask. As for prompt generation, we also start with sampling the real-world user cluster and obtaining its description of the typical users. Substantially, the ground truth prompt targeted at such user group is generated by GPT-4o. These modular and systematic designs enable G-MLLM to effectively incorporate group-level characteristics and preferences, enhancing its capability to generate contextually appropriate advertising backgrounds.

\subsection*{A.4 Details of Group Aggregation Algorithm}
\addcontentsline{toc}{subsection}{A.4 Details of Group Aggregation Algorithm}

In this section, we provide the detailed algorithmic description for the group  aggregation process outlined in the main paper. The algorithm demonstrates how we cluster users based on their product-aware preference representations and construct comprehensive group embeddings that capture both core and peripheral preferences within each cluster. The complete procedure is presented in Algorithm~\ref{alg:group_clustering}.

\begin{algorithm}[h!]
\small
\SetAlgoLined
\caption{Group Feature Aggregation}
\label{alg:group_clustering}
\SetKwInput{KwInput}{Input}
\SetKwInput{KwOutput}{Output}
\KwInput{%
    Trained preference model \\
    Product $s$ and its user set $\mathcal{U}_s$ \\
    $K_{\min}, K_{\max}$ — range for the number of clusters \\
    Percentile set $\mathcal{J}$ for representative point sampling
}
\KwOutput{%
    Group representations $\{\mathbf{G}_{s,1}, \mathbf{G}_{s,2}, \ldots, \mathbf{G}_{s,K_s^*}\}$
}
\BlankLine
\tcc{Step 1: Extract product-aware user embeddings}
$\mathcal{E}_s \leftarrow \{\mathbf{e}_{u|t} \mid u \in \mathcal{U}_s\}$\;
\BlankLine
\tcc{Step 2: Determine optimal number of clusters}
$S^* \leftarrow -\infty$, $K_s^* \leftarrow K_{\min}$\;
\For{$K = K_{\min}$ \KwTo $K_{\max}$}{
    $\{C_1, C_2, \ldots, C_K\} \leftarrow \text{KMeans}(\mathcal{E}_s, K)$\;
    $S_K \leftarrow \text{MeanSilhouetteCoefficient}(\mathcal{E}_s, \{C_1, \ldots, C_K\})$\;
    \If{$S_K > S^*$}{
        $S^* \leftarrow S_K$, $K_s^* \leftarrow K$\;
    }
}
\BlankLine
\tcc{Step 3: Final clustering and group representation construction}
$\{C_1, C_2, \ldots, C_{K_s^*}\} \leftarrow \text{KMeans}(\mathcal{E}_s, K_s^*)$\;
\For{$k = 1$ \KwTo $K_s^*$}{
    Compute cluster centroid $\mathbf{c}_k = \frac{1}{|C_k|}\sum_{\mathbf{e}_{u|t} \in C_k} \mathbf{e}_{u|t}$\;
    Sample peripheral points $\{\mathbf{p}_{k}^{(j)} \mid j \in \mathcal{J}\}$ at ${j}$ percentile distances from $\mathbf{c}_k$\;
    $\mathbf{G}_{s,k} \leftarrow \{\mathbf{c}_k\} \cup \{\mathbf{p}_{k}^{(j)} \mid j \in \mathcal{J}\}$\;
}
\Return $\{\mathbf{G}_{s,1}, \mathbf{G}_{s,2}, \ldots, \mathbf{G}_{s,K_s^*}\}$\;
\end{algorithm}

\subsection*{A.5 Limitations and Future Work}
\addcontentsline{toc}{subsection}{A.5 Limitations and Future Work}
In this section, we discuss some of the limitations of our current approach and outline potential directions for future improvements. 
One challenge we face is the relatively high computational cost of our method, which may hinder real-time deployment in dynamic advertising scenarios. 
While our approach provides highly personalized recommendations, its latency could affect its ability to deliver timely results across different user groups, limiting its real-world applicability. 
To address this, future work will focus on exploring more efficient architectures and optimization strategies to accelerate the workflow. By improving the processing speed, we aim to make our method more scalable and better suited for real-time personalized recommendations in fast-paced advertising environments.
In addition to improving efficiency, we also plan to enhance our method’s ability to generate images that incorporate textual elements, such as product descriptions, promotional messages, or brand slogans. 
This will further enrich the generated content and allow for more versatile and informative ad creatives, tailored to marketing objectives and campaign contexts.

\subsection*{A.6 Social Impact}
\addcontentsline{toc}{subsection}{A.6 Social Impact}

In this section, we discuss the broader social implications of automatic advertisement image generation. While such systems can boost efficiency and enable more tailored visual content, they may also amplify existing biases or lead to over-personalization that narrows users' exposure to diverse products and styles.
Our framework is designed to mitigate these risks by emphasizing fairness and inclusivity.
Group-level preference modeling helps avoid overfitting to niche groups, and the generation pipeline incorporates safety filters (e.g., official Stable Diffusion safety checkers) to block clearly inappropriate content. 
The system serves as an assistive tool for designers and marketers rather than a full replacement, aiming to enhance human creativity and productivity while maintaining human oversight and interpretability throughout real-world deployments.

\subsection*{A.7 Ethical and Legal Aspects}
\addcontentsline{toc}{subsection}{A.7 Ethical and Legal Aspects}
In this section, we clarify the ethical and legal principles governing user data handling and dataset usage in our work.
From an ethical perspective, we explicitly avoid targeting identifiable individuals. 
User information is processed entirely at the group level, with data anonymized and aggregated to ensure that no individual user can be re-identified.
Sensitive attributes are not used as explicit conditioning signals, and all outputs are checked to ensure the absence of personal portraits or other privacy-sensitive elements. AI-generated images are clearly labeled to support transparency and accountability.
On the legal side, all product images in our GAIP dataset are either owned by the partner platform or used under appropriate licenses that permit academic research. A dedicated legal and compliance team reviewed the data collection and usage protocol to confirm that it aligns with platform terms of service and applicable intellectual property regulations. The dataset is used solely for research and evaluation purposes, and no commercial deployment is conducted within the scope of this work.

\begin{figure*}[h]
   \centering
\includegraphics[width=\textwidth]{images/appendix_same_sku_2.pdf}
   \caption{More visualization examples generated by our method for the same product across different user groups. $\mathbf{G}_{*,k}$ denotes the $k$-th group feature for product ``$*$'', indicating that each column corresponds to a distinct group.}
   \label{fig:same_sku}
\end{figure*}

\begin{table*}[h]
\centering
\renewcommand{\arraystretch}{1.25}
\begin{tabular}{>{\centering\arraybackslash}m{1.9cm}|>{\centering\arraybackslash}m{1.9cm}|p{0.7\textwidth}}
\hline
\multicolumn{2}{c|}{\textbf{G-MLLM Task}} & \multicolumn{1}{c}{\textbf{Instructions}} \\
\hline

\multirow{8}{*}[-110pt]{\shortstack{Group \\ Centric}}
& \multirow{4}{*}[-50pt]{\shortstack{Group \\ Analysis}}
    & \textit{Regarding consumer \textless{}users\textgreater{} on the e-commerce platform, please describe the characteristics of its most common users.} \\ 
  &
    & \textit{For this type of consumer \textless{}users\textgreater{}, please analyze its most common characteristics. You need to focus on member\_level, promotions\_sensitive\_level, purchase\_power, gender, marriage, customer\_points, is\_vip, is\_impulse\_purchase\_user, per\_month\_order\_num, os\_plant, has\_car, has\_house, resident\_city\_level, child\_age, profession, user\_active, age, target\_group, education, color.} \\ 
  & 
    & \textit{There is a type of consumer \textless{}users\textgreater{}. What kind of person do you think he is? Consider these aspects, such as member\_level, promotions\_sensitive\_level, purchase\_power, gender, marriage, customer\_points, is\_vip, is\_impulse\_purchase\_user, per\_month\_order\_num, os\_plant, has\_car, has\_house, resident\_city\_level, child\_age, profession, user\_active, age, target\_group, education, color.} \\ 
  & 
    & \textit{You are an expert who specializes in studying e-commerce consumer portraits. Now there is a type of consumer whose characteristic is \textless{}users\textgreater{}. Please describe his portrait based on the needs of the e-commerce field.} \\ \cline{2-3}

  & \multirow{4}{*}[-50pt]{\shortstack{Behavioral \\ Prediction}}
    & \textit{The item titled \{\} has been historically clicked by consumers \textless{}users\textgreater{}. Can you predict the characteristics of another possible user who will click this item? You just need to pay attention to member\_level, promotions\_sensitive\_level, purchase\_power, gender, marriage, customer\_points, is\_vip, is\_impulse\_purchase\_user, per\_month\_order\_num, os\_plant, has\_car, has\_house, resident\_city\_level, child\_age, profession, user\_active, age, target\_group, education, color.} \\ 
  & 
    & \textit{After clicked by these consumers \textless{}users\textgreater{}, who is the next consumer that may be keen on the item titled \{\}? Please describe his characteristic.} \\ 
  & 
    & \textit{Considering the fact that several consumers \textless{}users\textgreater{} have clicked the same item \{\}, forecast who is the next user that will be interested in this item.} \\ 
  & 
    & \textit{You have access to the historical user interaction record \textless{}users\textgreater{} of the item \{\}. Now your task is to predict the feature of the another possible user that loves the same item based on the past interaction.} \\
\hline

\multirow{8}{*}[-50pt]{\shortstack{Product \\ Centric}}
  & \multirow{4}{*}{\shortstack{Product \\ Understanding}}
    & \textit{\textless{}image\textgreater{} What words best capture the product presented in this image?} \\ 
  & 
    & \textit{\textless{}image\textgreater{} Summarize the product in this image with a caption that captures its essence.} \\ 
  & 
    & \textit{\textless{}image\textgreater{} Construct a succinct yet descriptive caption for the displayed product image.} \\ 
  & 
    & \textit{\textless{}image\textgreater{} Author a caption for the product image that summarizes its attributes.} \\ \cline{2-3}

  & \multirow{4}{*}[-50pt]{\shortstack{Prompt \\ Generation}}
    & \textit{Design a concise Stable Diffusion prompt that takes the product caption \{\} and product image as inspiration to generate an appealing advertising image background for this product. Your answer should be targeted at consumer \textless{}users\textgreater{}.} \\ 
  & 
    & \textit{Generate a short Stable Diffusion prompt that leverage the product caption \{\} and the visual elements of the product image to output an advertising background that underscores the product's attractiveness. Please consider the characteristics of consumer \textless{}users\textgreater{} and answer.} \\ 
  & 
    & \textit{Develop a compact prompt for Stable Diffusion to craft a background for an ad, using the product caption \{\} and the accompanying product image as creative influences. Consumers with characteristic \textless{}users\textgreater{} is your target group.} \\ 
  & 
    & \textit{Draft a distilled text to image prompt to fabricate an ad background, infusing the essence of \{\} from the product caption and the picture to accentuate the product's features. Please pay attention to the consumer with characteristic \textless{}users\textgreater{} and make the most suitable choice for him.} \\
\hline
\end{tabular}
\caption{Instruct directives for Group-aware Multimodal Large Language Model.}
\label{tab1}
\end{table*}